# BagBERT: BERT-based bagging-stacking for multi-topic classification

Loïc Rakotoson, Charles Letaillieur, Sylvain Massip and Fréjus Laleye
{firstname.lastname}@opscidia.com
Opscidia, Paris, France

*Abstract*— **This paper describes our submission on the COVID-19 literature annotation task at Biocreative VII. We proposed an approach that exploits the knowledge of the globally non-optimal weights, usually rejected, to build a rich representation of each label. Our proposed approach consists of two stages: (1) A bagging of various initializations of the training data that features weakly trained weights, (2) A stacking of heterogeneous vocabulary models based on BERT and RoBERTa Embeddings. The aggregation of these weak insights performs better than a classical globally efficient model. The purpose is the distillation of the richness of knowledge to a simpler and lighter model. Our system obtains an Instance-based F1 of 92.96 and a Label-based micro-F1 of 91.35.**

## I. INTRODUCTION

Our submitted systems consist of several knowledge aggregation techniques using ensembles models. Each single model performs end-to-end multilabel classification of predefined topics on scientific publications related to COVID-19 in the health sector. The inputs to the model are the various fields of a document, including in our case the title, keywords as well as the abstract, and are returned none, one or more topics related to the document. Our systems use BERT (1) and WordNet (2) for initializing the training data, and the Encoder part of the Transformers architecture (3) for modeling.

As each of our submissions is an evolution of the previous one, we first describe how we processed the training data and then talk about the routing to the final system which is a combination of bagging and stacking.

## II. DATASETS

The raw training data are provided by the LitCovid database (4,5) which are used for the specific topic multilabel classification task for Biocreative VII. It contains about 25k rows with unbalanced labels, i.e. a ratio of 15 between majority and minority labels, and groups of labels that are represented only once.

This strong imbalance is a risk of overfitting. Following our first experiments which revealed the poor predictive performance, specifically for the minority class, we finally turned towards the ensemble methods, in particular the bagging approach. In the current state of the data, a bootstrap resampling will lower the representativeness of each sample due to the groups of labels that appear only once. We have chosen to neglect the independence of the samples and to focus on the representativeness. The initial training dataset will then have several versions with the combinations of data augmentation.

### A. Field Order

We consider only the title, the keywords and the abstract as carrying main information about the document and we take only the first 350 tokens of their concatenation. Thus, some information at the end is omitted. The information carried by these fields have different importance for each label, and the same is true for each part of the abstract. So, we have built a version with the title in front and the keywords sometimes missing from the dataset, that are neglected, and a second version that highlights these keywords and omits the abstract conclusion.

### B. Nominal groups masking

The collection of documents is all about COVID-19 so, we have built a dataset version that then considers all the terms related to it as a constant between all the labels, thus stopwords to be deleted or masked at the expense of the text structure.

### C. Substitution and a noise addition

To extend the vocabulary we randomly substituted tokens with their synonyms using WordNet and others with a token or group of tokens that are contextually close to them using a zero-shot BERT. This augmentation also aims at improving the robustness of our models.

This results in 8 marginally dependent samples of training data of about 15k rows each, due to the weighted subsampling of the dominant label groups.

## III. SYSTEM CONSTRUCTION

### A. Method

We based our models on BERT (1) and RoBERTa (6) with different training corpora and thus with different vocabularies and language models. We used 3 BERT models with vocabularies from PubMed (7), on clinic data (8) and SciBERT on CORD-19 (9,10). On the other hand we have an agnostic RoBERTa in its base version and a second one trained on scientific articles from Semantic Scholar (11) named BioMed. We optimize our models during training with respect to binary cross entropy and Hamming loss and observe the F1 score for labels and instances.

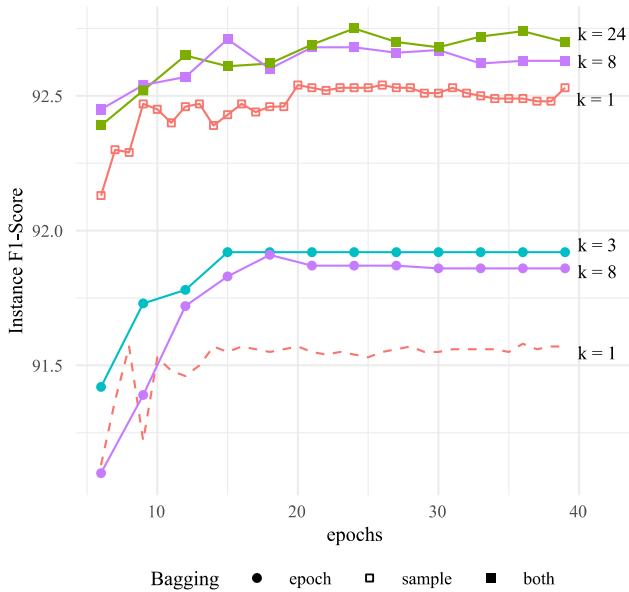

Fig. 1. Evolution of the F1 score of PubMedBERT models as related to the type of bagging and the number k of epochs taken in the ensemble model. Performance without any bagging is in red dotted line with $k=1$. Performance with a classical bagging is in red and square shape with $k=1$.

### B. Model Bagging

*1) Epochs bagging*

For a single model, instance-based metrics improve over epochs while label-based metrics are less stable and oscillate. This is due to the imbalance of the data which, even if reduced, produces overfitting. The weights computed on the epochs prior to the optimal weight according to the F1 on the instance give better performances on the underrepresented labels. So, we consider the models built from the $k$ weights close to the optimal one according to the Hamming distance as weak learners whose aggregation allows a more correct representation. With an aggregation of $k$-weights of 3 and 8 on the bagging of epochs (Fig. 1), the predictions outperform the classical learning ($k=1$) after 10 epochs and obtain the optimal ensemble before the 20th step, while this stage is fully reached only 15 epochs later for the classical model.

*2) Sample bagging*

The classical bagging of the model on the 8 samples allows it to be more robust thanks to the variance due to the noise introduced in the data. We have combined the two bagging by taking the $k =$ top-$n$ best epochs. By taking the top-3 of the 8 samples ($k=24$), we obtain very good performance, however by making a less complex set by taking the 8 best of the top-3 of each sample ($k=8$) the performances are approximately the same (Fig. 1).

We made an initial submission of the best bagging model which is PubMedBert (TABLE 1).

TABLE 1. *k*-3 Bagging Models labels F1 score

|  | PubmedBert | SciBert | BioMed | BioClinical |
|---|---|---|---|---|
| Treatment | 90.08 | 90.25 | **90.78** | 89.58 |
| Diagnosis | 87.82 | 87.07 | 86.73 | **89.14** |
| Prevention | **94.68** | 94.09 | 93.99 | 93.97 |
| Mechanism | **88.56** | 87.38 | 87.25 | 88.22 |
| Transmission | **71.46** | 69.11 | 69.4 | 67.84 |
| Forecasting | 76.13 | **78.15** | 70.6 | 74.6 |
| Case Report | **92.34** | 90.85 | 90.2 | 90.03 |

The best *k*-3 bagging models for each label in terms of F1 score. Only the agnostic RoBERTa is not included.

### C. Model Stacking

The created bagging models are homogeneous ensembles according to their vocabulary, both by the way of tokenization and the weights of the input tokens. Our goal is to combine several different representations as well as the specializations of each in the prediction of labels (TABLE 1).

Unlike classical stacking, the aggregated models are not the averaged models. Indeed, we have extended the epoch bagging strategy on all the target models. The ensemble method at this stage then consists in building a meta-model by selecting the top-$k$ epochs of the $n$ models and averaging over the $nk$ epochs.

We submitted two MetaRoberta-$k$ with $k \in \{1, 3\}$ using this strategy, i.e. two ensembles with Biomed and Roberta agnostic. Looking at the complexity/score ratio, the performance increase is negligible while the complexity is multiplied by $k$ (TABLE 2).

The MetaBert submission weights $k$ according to the Hamming loss to improve the score without exponentially increasing the complexity. This method parsimoniously selects the $k$-epochs to compose the meta-model: a lower loss gives a higher $k$ value, with $k \subset [2, 4]$.

TABLE 2. Submission performance

|  | Label based F1 | | Instance F1 |
|---|---|---|---|
|  | *macro* | *micro* |  |
| PubmedBert | 87.45 | 90.89 | 92.61 |
| MetaRoberta-1 | 87.28 | 90.87 | 92.37 |
| MetaRoberta-3 | 87.42 | 90.88 | 92.41 |
| MetaBert | 87.43 | 90.99 | 92.70 |
| MetaEnsemble | **88.24** | **91.35** | **92.96** |
| ML-NET* | 76.55 | 84.37 | 86.78 |
| Track Q3* | 86.70 | 90.83 | 92.54 |

Results of our submissions on test data with a bagging model and 4 meta-models. The richest representation MetaEnsemble concentrates the best scores. (*track-5 submissions statistics)

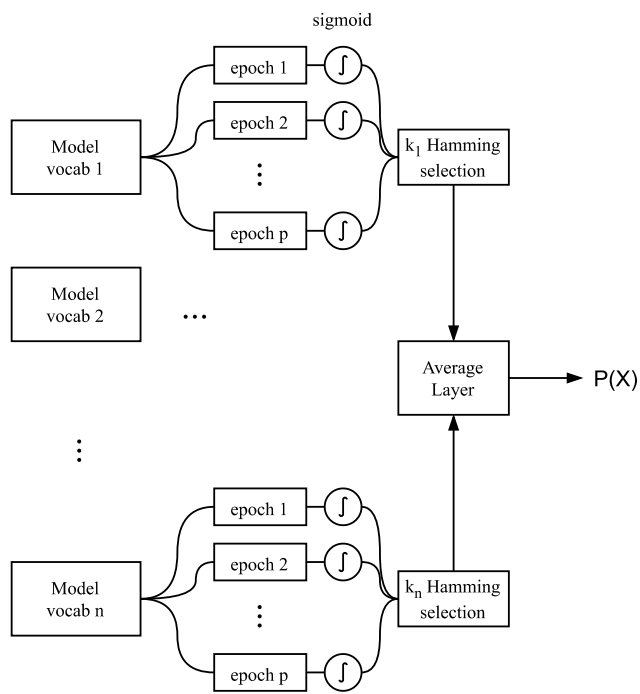

Fig. 2. MetaEnsemble system architecture with parallel and independent model training

Finally, we submitted a MetaEnsemble (Fig. 2) which consists of combining all models with the weighting of k to have the richest representation achievable. We also report the performance of ML-NET as a baseline as well as the Q3 of submissions for this task (12) .

*D. Ensemble Purpose*

The MetaEnsemble is usable as is since it offers the best performance. However, its purpose is to distill its knowledge into a much lighter and less complex model.

A model that is not deep enough will have trouble capturing complex relationships without risking overfitting from the binary values of each label alone. The goal of MetaEnsemble distillation is to predigest the modeling so that a small model directly accesses the richest representation.

With the construction of the MetaEnsemble representation from the data samples for bagging, in addition to being able to generalize quickly, the student-model will:

- be able to build the correct representation despite the lack of information such as keywords or complete parts of texts during its inference,
- have an effective knowledge of the terms related to the COVID-19 when it is needed even if it is not explicit,
- be able to easily represent tokens that are not found in the vocabularies of the current articles but are related to them with good robustness.

There are several candidates for the student-model, which leaves the choice free to the end-user according to his resources as the performance is already present.

IV. CONCLUSION

This paper describes our submissions for track 5 of BioCreative VII. We present the system we developed which consists of a combination of epoch bagging and stacking of BERT-based models of different vocabularies by training them on augmented or noisy data. This system performs 92.96 of F1 based on the instance with 91.35 of macro-F1 on the labels. The final system to be developed is its distillation according to the resources available for its use. The code will be available in open source
(at https://github.com/opscidia/Bagbert).